\documentclass{article}
\pdfoutput=1

     \usepackage[final]{neurips_2022}

\usepackage{microtype}
\usepackage{graphicx}
\usepackage{subfigure}
\usepackage{amsmath}
\usepackage{amssymb}
\usepackage{booktabs} % for professional tables
\usepackage{color, colortbl}
\usepackage{xcolor}         % colors
\definecolor{Gray}{gray}{0.93}

\usepackage{hyperref}

\title{Learning Regularized Positional Encoding\\for Molecular Prediction}

\author{%
    Xiang Gao, Weihao Gao, Wenzhi Xiao, Zhirui Wang, Chong Wang\thanks{Currently affiliated with Apple Inc., work done at ByteDance Inc.}, Liang Xiang \\
  ByteDance Inc. \\
  \\
  \texttt{\{xianggao,weihao.gao,xiaowenzhi\}@bytedance.com} \\
  \texttt{\{zhirui.wang,xiangliang\}@bytedance.com,mr.chongwang@gmail.com} \\
}

\begin{document}

\maketitle

\begin{abstract}

Machine learning has become a promising approach for molecular modeling. Positional quantities, such as interatomic distances and bond angles, play a crucial role in molecule physics. The existing works rely on careful manual design of their representation. 
To model the complex nonlinearity in predicting molecular properties in an more end-to-end approach, we propose to encode the positional quantities with a learnable embedding that is continuous and differentiable.
A regularization technique is employed to encourage embedding smoothness along the physical dimension. 
We experiment with a variety of molecular property and force field prediction tasks. Improved performance is observed for three different model architectures after plugging in the proposed positional encoding method.
In addition, the learned positional encoding allows easier physics-based interpretation. We observe that tasks of similar physics have the similar learned positional encoding.%\footnote{The code will be open-sourced on GitHub.}

\end{abstract}
\section{Introduction}
\label{sec-intro}

The prediction of molecular properties is crucial for the discovery of molecules with desired properties, with applications in drug, material, and energy industries. 
\emph{Ab initio} quantum chemical calculations, such as these based on density functional theory (DFT), give accurate predictions for the chemical properties with a high computational cost.
Semi-empirical methods \cite{stewart2007pm6method} are faster but much less accurate.
Machine learning based molecular predictor is a promising alternative solution to balance accuracy and efficiency. 
In recent years scientists have started leveraging machine learning methods in molecular prediction tasks \cite{rupp2012fast, montavon2013machine}.

An effective machine learning model should represent the key physical quantities properly. 
Positional quantities, such as interatomic distances and bond angles, are among the most fundamental physical quantities for ML-based molecular predictors.
According to Born-Oppenheimer approximation, a molecular system is uniquely determined by the nucleus positions and charges. Almost all molecular properties are thus a function of the positions and types of the atoms.
%\emph{(i)} Positional quantities are directly used in the definition of properties of interest, such as the distance used in dipole moment.
%\emph{(ii)} 
Moreover, interatomic distances and bond angles are invariant to transformations such as rotations and translations. These symmetric properties make them preferred inputs to build machine learning models to exploit the symmetry of physical problems \cite{miller2020l1net, finzi2020lieconv, satorras2021egnn}.
%\emph{(iii)} Many important physical quantities strongly depend on positional quantities. The change of distance reflects the change of dominant mechanisms. For instance, in the case described by Lennard-Jones potential, two atoms are repulsive at short distance, but attractive at long distance. More generally, the interatomic potential energies are often a function of positional quantities and other variables. 

The design of an effective positional quantity representation faces challenges from the complex physics in molecules.
First, the interactions in molecules are often nonlinear or even non-monotonic with respect to the positional quantities. Lennard-Jones potential, for instance, is a function of a repulsive term and an attractive term. %This nonlinear function is also non-monotonic, with the repulsive term being dominant the short-range and the attractive on the long-range. 
%It is challenging for a machine learning model to learn these nonlinear transformation from the scratch.
Second, multiple nonlinear effects co-exist and they scales differently with positional quantities. For example, Lennard-Jones potential involves two polynomial terms, while Morse potential contains two exponential terms. 
%Ideally, a variety of nonlinear transformation should be employed simultaneously to press these complex dependence.
Ideally, a good representation of positional quantities should help the machine learning models to effectively model these multi-modal non-linearity.

Existing methods of positional quantities representations can be grouped into three categories. 
(\emph{i}) Directly using original scalar form as the model inputs \cite{satorras2021egnn}. Although neural networks can learn any function of the inputs according to the universal approximation theorem, it is practically challenging given the multiple complex physics in molecules mentioned above.
(\emph{ii}) Using a manually defined set of simple transformations, such as the reciprocal of interatomic distances used by Wang et.al.~\citep{wang2018deepmd}. Such representations maybe more effective than directly using original scalar forms, but they are only ideal for limited potential functions. For example, Lennard-Jones potential contains polynomial terms of reciprocal of interatomic distances but Morse potential do not. The generalizability of such a method is limited.
(\emph{iii}) Using the representation of positional quantities under a set of manually picked basis functions , such as the Gaussian kernels \cite{bartok2017machine, chmiela2017machine} or Bessel functions \cite{klicpera2020dimenet}. This method also has limited generalizability and the efficiency depends on the choice of the basis functions. 

Motivated by the success of learnable positional encoding in natural language processing (NLP) \cite{devlin2018bert, brown2020gpt3}, we propose to use learnable positional encoding in molecular properties prediction.%, as illustrated in Figure~\ref{fig:overview}.
Compared to the existing methods, the learnable position embedding is not limited to a manually picked function format. Empirically we demonstrate that our method improve the accuracy on multiple molecular property and force field tasks compared to state-of-the-art models.
Our method works as a plug-in module, combined with any models with positional quantities as inputs.
Moreover, we propose a smoothness regularization technique, which filters out unnecessary embedding fluctuations and makes physics-based interpretation possible. We demonstrate that the learned embedding reflects the dependence of different tasks on short and long-range interactions, and observe that tasks of similar physical nature result in similar positional encoding.

Our contribution is two-folds. 

1. We propose a learnable positional encoding method that is continuous and differentiable. The method can be used to represent interatomic distance, bond angles and other positional quantities with a series of nonlinear transformations. Experiments show that new state-of-the-art results are achieved with this technique.

2. We propose a regularization method for the proposed positional encoding. Based on physical intuition, this method encourages the embedding to form a meaningful manifold for easier visualization and interpretation from a physical point of view. The regularization reduces overfitting and can further increase model accuracy for certain tasks.

\section{Related works}

\subsection{Positional encoding}
There are generally two types of positional encoding, fixed and learned. Fixed positional encoding represent positions with manually picked basis functions. In the initial version of transformer \cite{vaswani2017transformer}, the discrete token positions are represented by a series of sine and cosine functions of a range of fixed frequencies. Gaussian kernels are employed to represent some continuous physical quantities such as interatomic distances in works of molecule modeling \cite{bartok2017machine, chmiela2017machine}. DimeNet \cite{klicpera2020dimenet, klicpera2020dimenetpp} and GemNet~\cite{klicpera2021gemnet} and SphereNet~\cite{liu2021spherical} instead use Spherical Bessel functions and spherical harmonics. Learned positional encoding represent each position with a learned vector. Although initially the fixed positional encoding was proposed for transformers \cite{vaswani2017transformer}, the learnable positional encoding has become more popular technique in natural language processing (NLP) community, as in BERT \cite{devlin2018bert} and GPT models \cite{radford2018gpt1, radford2019gpt2, brown2020gpt3}.
Besides transformers, learnable positional encoding is used with convolutional neural networks \cite{gehring2017convolutional}.

%In contrast to the molecular prediction problems in this work, the previous application of the learned positional encoding are designed for discrete tokens. That technique cannot be directly applied to the continuous positions for molecular modeling.

\subsection{Molecular modeling}

Convolution neural networks are equivariant to translations and this enables them to generalize better on image problems. This characteristic inspires researcher to use them in molecular properties prediction, where translation equivalence is a useful inductive bias \cite{schutt2017schnet, miller2020l1net, finzi2020lieconv}.
Graph neural networks are another popular family of models used in modeling molecules. \cite{gilmer2017nmp, satorras2021egnn, fuchs2020se,dwivedi2021graph, klicpera2020dimenetpp, klicpera2020dimenet}.
%There are promising efforts to employ other model architectures in the molecular modeling problems. Anderson et.al.\cite{anderson2019cormorant} proposed a Cormorant network based on neurons designed to be rotationally invariant and takes angular feature into consideration. Ying et.al.
%\cite{ying2021graphormer} applied transformers on graph data. They added useful graph features as additional features for token embedding, and modified the attention mechanism considering molecule mechanics.

%The regularized positional encoding method proposed in the present work is a technique that are generally not limited by the choice of neural network architectures reviewd above. As a plug-in module, researchers can replace the raw positional input by the proposed embedding.

\section{Method}
\label{sec-method}

In this section, we introduce our method to encode positional quantities such as interatomic distance or bond angle, to a fixed-length learnable vector, $h(x) \in \mathbb{R}^s$, where $s$ is the embedding size.

\subsection{Continuous and Differentiable Positional Encoding}

In contrast to the application in NLP, where the positional embedding are designed for discrete tokens, the positional quantities are continuous for molecule modeling.
To represent these \emph{continuous} physical quantities with embeddings on \emph{discrete} points, we divide the space of $x$ into $n_\text{bin}$ bins, and obtain the embedding of a arbitrary $x$ by interpolating the embeddings of the nearest bin centers. 

A simple implementation is via linear interpolation 
\begin{align*}
     h(x) & = (1 - t) h(\lfloor x \rfloor) + t h (\lceil x \rceil) , \quad t \equiv \frac{x - \lfloor x \rfloor} {\lceil x \rceil - \lfloor x \rfloor} ,
\end{align*}
where $\lfloor x \rfloor$ is the largest bin center smaller than $x$, and $\lceil x \rceil$ is the smallest bin center larger than $x$. 

%This implementation involves a sets of learnable parameters, $\{ h(x_i) \}$, where $i = 1,2,...,n_\text{bin} $ is the bin index, and $\{ x_i \}$ are the bin centers. The total number of additional trainable parameters is $s n_\text{bin}$

The linear interpolation method is simple but it makes $h(x)$ not differentiable with respect to~$x$, as the derivative $g(x) \equiv \frac{d h(x)}{d x} \in \mathbb{R}^s$ is not continuous at the bin centers.
% \begin{align*}
%     g(x) \equiv \frac{d h(x)}{d x} \in \mathbb{R}^s,
% \end{align*}
$g(x)$ is required if the trained model is used in calculation that involving the derivative with respect to~$x$. For example, the potential force acting on atoms is the derivative of potential energy with respect to their positions. 
%In such case, the distance embedding should be differentiable w.r.t. distance. This requires the following derivative $g$ to exist and be continuous.
%The linear interpolation method as defined in Equation \ref{eq:linear} does not satisfy this requirement, as $g$ is not continuous at the bin centers. 
To resolve this issue, we consider higher order interpolation. We choose the Cubic Hermite spline as follows,
\begin{align*}
     h(x) & = c_1 h(\lfloor x \rfloor) + c_2 h (\lceil x \rceil) + c_3 g(\lfloor x \rfloor) + c_4 g (\lceil x \rceil),
\end{align*}
where the interpolation coefficients are
\begin{align*}
    c_1 &= 2 t^3 - 3 t^2 + 1, \, c_2 = 1 - c_1, \, c_3 = t^3 - 2 t^2 + t, \, c_4 = t^3 - t^2.
\end{align*}
The positional encoding $h(x)$ defined in this way is continuous and differentiable with respect to~$x$. 

This implementation involves two sets of learnable parameters, $\{ h(x_i) \}$ and $\{ g(x_i) \}$, where $i = 1,2,...,n_\text{bin} $ is the bin index, and $\{ x_i \}$ are the bin centers. The total number of additional trainable parameters is $2 s n_\text{bin}$. The proposed encoding method provides a simple way to build a universal approximation of any continuous functions of $x$. The accuracy of the approximation can be easily increased by adding the number of bins.

\subsection{Smoothness Regularization}

As the positional quantities are continuous, we assume the embedding should be locally smooth. The difference between the embedding of two adjacent bins should not be large. Following this intuition we consider a smoothness loss, defined as the average relative change of a embedding, $h(x_{i})$, compared to the next bin, $h(x_{i+1})$.
\begin{align*}
     \mathcal{L}_\text{smooth} &= \frac{ \sum_{i=1}^{n-1}{|| h(x_{i+1}) - h(x_{i}) ||}}{ \sum_{i=1}^{n-1}{ || h(x_i) ||}}.
\end{align*}
where $||.||$ is 2-norm. During training, $\mathcal{L}_\text{smooth}$ is combined with the original training loss $\mathcal{L}_\text{orig}$ as the new training loss
\begin{align*}
     \mathcal{L} &=  \mathcal{L}_\text{orig}  + \lambda \mathcal{L}_\text{smooth} ,
\end{align*}
where $\lambda$ is a hyperparameter to control the contribution of the smoothness loss.

$\mathcal{L}_\text{smooth}$ acts as a regularization term as it reduces the flexibility of the positional encoding. This may help the model generalize in case of small amount of data. Even if a bin is never directly trained during training (i.e., no $x$ fall near this bin during training), its parameters are still updated because its difference with neighbor bins is regularized by the smoothness loss. We will show that smoothness regularization can improve model accuracy and provide physical interpretability in the next section.

\subsection{Implementation}
%\subsection{Plug into Existing Models}
\label{sec:implement}

The regularized positional encoding is proposed as a plug-in for existing models. When a model uses positional quantities as their input, one may consider replacing these quantities by the proposed embedding, as illustrated in Figure~\ref{fig:overview}. In this work, we consider three backbond models:  EGNN \cite{satorras2021egnn}, DimeNet++ \cite{klicpera2020dimenetpp}, and a Transformer\cite{vaswani2017transformer} based model\footnote{It uses self attention mechanism to model the interaction among atoms. The interatomic distance is sent to a multi-layer perceptron (MLP) as the bias vector for the multi-head attention}. The experiments are conducted on a Nvidia Tesla V100 GPU.

\section{Experiments and discussion}
\label{sec-results}

\subsection{Molecular force fields}

\begin{table}[!ht]
    \centering
    \small
    %\footnotesize
    %\scriptsize

    \renewcommand{\arraystretch}{1.1}

    \begin{tabular} {p{0.12\textwidth} | 
    p{0.06\textwidth} p{0.06\textwidth} p{0.06\textwidth} p{0.12\textwidth} p{0.09\textwidth} p{0.07\textwidth} p{0.07\textwidth} p{0.07\textwidth}}
     
    \hline
    & Aspirin & Benzene & Ethanol & Malonaldehyde & Naphthalene & Salicylic & Toluene & Uracil \\
    \hline
    sGDML & 29.5 & 2.6 & 14.3 & 17.8 & 4.8 & 12.1 & 6.1 & 10.4 \\
    FCHL19 & 20.7 & - & 5.9 & 10.6 & 6.5 & 9.6 & 8.8 & 4.6 \\
    DimeNet & 21.6 & - & 10.0 & 16.6 & 9.3 & 16.2 & 9.4 & 13.1 \\
    SphereNet & 18.6 & - & 9.0 & 14.7 & 7.7 & 15.6 & 6.7 & 11.6 \\
    NequIP & 15.1 & 2.3 & 9.0 & 14.6 & 4.2 & 10.3 & 4.4 & 7.5\\
    PaiNN & 14.7 & - & 9.7 & 14.9 & \textbf{3.3} & 8.5 & \textbf{4.1} & \textbf{6.0}\\
    \hline
    Transformer & 49.7$_{(0.9)}$ & 36.8$_{(0.6)}$ & 51.0$_{(1.3)}$ & 53.0$_{(1.5)}$ & 50.4$_{(1.5)}$ & 53.2$_{(1.6)}$ & 47.6$_{(1.5)}$ & 54.5$_{(1.9)}$ \\
    \rowcolor{Gray} +PosEnc & 24.6$_{(0.8)}$ & 3.3$_{(0.2)}$ & 13.8$_{(0.3)}$ & 24.2$_{(0.7)}$ & 11.6$_{(0.5)}$ & 22.5$_{(0.9)}$ & 15.2$_{(0.5)}$ & 17.7$_{(0.8)}$\\
    \rowcolor{Gray} ~~~+Smooth & 12.3$_{(0.5)}$ & 1.9$_{(0.2)}$ & 6.2$_{(0.5)}$ & 13.8$_{(0.8)}$ & 5.4$_{(0.4)}$ & 8.3$_{(0.7)}$ & 5.7$_{(0.2)}$ & 10.1$_{(0.2)}$\\
    \hline
    EGNN & 51.9$_{(1.1)}$ & 36.9$_{(0.4)}$ & 50.1$_{(1.3)}$ & 53.5$_{(1.8)}$ & 50.5$_{(1.2)}$ & 53.3$_{(0.9)}$ & 47.6$_{(0.4)}$ & 54.6$_{(0.6)}$ \\
    \rowcolor{Gray} +PosEnc & 9.3$_{(0.8)}$ & \textbf{1.5}$_{(0.2)}$ & 4.9$_{(0.3)}$ & 8.8$_{(0.7)}$ & 4.9$_{(0.4)}$ & 7.8$_{(0.8)}$ & 7.7$_{(0.5)}$ & 7.4$_{(0.6)}$ \\
    \rowcolor{Gray} ~~~+Smooth & \textbf{6.7}$_{(0.5)}$ & \textbf{1.3}$_{(0.2)}$ & \textbf{3.5}$_{(0.3)}$ & \textbf{7.0}$_{(0.8)}$ & \textbf{3.0}$_{(0.3)}$ & \textbf{6.5}$_{(0.5)}$ & \textbf{4.1}$_{(0.4)}$ & \textbf{6.2}$_{(0.4)}$ \\
    \hline
\end{tabular}
\vspace*{0.1in}
\caption{Test error (MAE in meV/Å) on MD17 DFT dataset. \colorbox{Gray}{Gray rows} are our methods. \textbf{Bold} text indicates the best error. The numbers in brackets are the standard deviation.}
\label{tab:ff_out}
\label{tab:ff}
\end{table}

We experiment with a set of molecular force field datasets, MD17 \cite{chmiela2017md17}. The task is to predict the force acting on each atom given the 3D geometry (configuration) of the molecule. Following \cite{batzner2022nequip}, we use 1000 training and test configurations.

Surprisingly, as illustrated in Table~\ref{tab:ff}, both EGNN and the Transformer-based model shows a large error, although these two architectures previously achieve success in a various tasks \cite{satorras2021egnn, ying2021graphormer}.
We hypothesise the data characteristics of the current task makes the relatively simple representation of the interatomic distance $r$ by EGNN and the Transformer-based model performing not well on this task.
Molecular force fields are highly sensitive to the change of 3D geometries. The 3D geometries of the same molecule in MD17 dataset only slightly different from each other, while the forces acting on atoms change significantly across molecular configurations. 

We then use the proposed positional encoding method (+PosEnc in Table~\ref{tab:ff}) to represent the interatomic distance. This significantly reduce the test error for both EGNN and the transformer-based model. 
When we employ the proposed smoothness regularization technique (+Smooth in Table~\ref{tab:ff}), the test error further decreases. 

We compare the results with a few recent works, sGDML \cite{chmiela2019sgdml}, FCHL19 \cite{christensen2020fchl}, DimeNet \cite{klicpera2020dimenet}, SphereNet \cite{coors2018spherenet}, NequIP \cite{batzner2022nequip}, and PaiNN \cite{schutt2021painn}. They either use kernels or carefully designed spatial basis functions to represent the positional quantities. We show that with the proposed learnable positional encoding technique, similar or better results on force fields can be achieved, as illustrated in Table~\ref{tab:ff}.

\subsection{Molecular properties}

\begin{table*}[!ht]
    \centering
    \small
    \begin{tabular}{p{0.095\textwidth} | p{0.03\textwidth} p{0.03\textwidth} p{0.03\textwidth} p{0.03\textwidth}  p{0.03\textwidth} p{0.09\textwidth} p{0.03\textwidth} p{0.03\textwidth} p{0.03\textwidth} p{0.03\textwidth} p{0.03\textwidth} p{0.05\textwidth}}   
    
    \hline
    
    Task & $\alpha$ & $\Delta\epsilon$ & $\epsilon_\text{HOMO}$ & $\epsilon_\text{LUMO}$ & $\mu$ & $C_v$ & $G$ & $H$ & $R^2$ & $U$ & $U_0$ & ZPVE \\
    Unit & bohr$^3$ & meV & meV & meV & D & cal/mol K & meV & meV & bohr$^3$ & meV & meV & meV \\%& meV \\
    \hline
    NMP       & .092 & 69 & 43 & 38 & .030 & .040 & 19 & 17 & .180 & 20 & 20 & 1.50  \\
    SchNet    & .235 & 63 & 41 & 34 & .033 & .033 & 14 & 14 & .073 & 19 & 14 & 1.70  \\
    Cormorant & .085 & 61 & 34 & 38 & .038 & .026 & 20 & 21 & .961 & 21 & 22 & 2.03  \\
    L1Net.    & .088 & 68 & 46 & 35 & .043 & .031 & 14 & 14 & .354 & 14 & 13 & 1.56  \\
    LieConv.  & .084 & 49 & 30 & 25 & .032 & .038 & 22 & 24 & .800 & 19 & 19 & 2.28  \\
    
    \hline
    DimeNet++ & .048 & 44 & 27 & 21 & .032 & .023 & 7 & 7 & .334 & 7 & 7 & 1.18 \\
    +MLP & .059 & 49 & 30 & 25 & .035 & .027 & 11 & 11 & .286 & 11 & 10 & 1.54 \\
    \rowcolor{Gray} +PosEnc & .048 & 44 & 26 & \textbf{20} & .030 & .023 & 7 & 7 & \textbf{.260} & 7 & 7 & 1.18 \\
    \rowcolor{Gray} ~~~+Smooth & .048 & 44 & \textbf{24} & \textbf{20} & \textbf{.028} & .023 & 7 & 7 & \textbf{.260} & 7 & 7 & 1.18 \\
    
    \hline
    EGNN & .071 & 48 & 29 & 25 & .029 & .031 & 12 & 12 & .106 & 12 & 11 & 1.55  \\
    +MLP & .068 & 45 & 35 & 24 & .027 & .029 & 10 & 10 & .122 & 11 & 10 & 1.50 \\
    \rowcolor{Gray} +PosEnc & .063 & 44 & \textbf{26} & \textbf{22} & .027 & \textbf{.028} & 10 & 10 & \textbf{.086} & 10 & \textbf{9} & \textbf{1.49} \\
    \rowcolor{Gray} ~~~+Smooth & \textbf{.062} & \textbf{43} & 27 & 23 & \textbf{.023} & \textbf{.028} & \textbf{9} & \textbf{9} & .091 & \textbf{9} & \textbf{9} & 1.50 \\
    
    \hline
    
    \end{tabular}
    \caption{Test error on QM9 dataset. \colorbox{Gray}{Gray rows} are our methods. \textbf{Bold} text indicates the best error of a given backbone model.}
    \label{table:eval}
\end{table*}

We then experiment with the molecular property prediction tasks using the QM9 \cite{ramakrishnan2014qm9} dataset. This dataset consists the DFT calculation results of 134 k stable small organic molecules. The geometries minimal in energy with a variety of corresponding chemical properties are included: isotropic polarizability ($\alpha$), energy of HOMO ($\epsilon_\text{HOMO}$), energy of LUMO ($\epsilon_\text{HOMO}$), energy gap ($\Delta \epsilon = \epsilon_\text{HOMO} - \epsilon_\text{HOMO}$), dipole moment ($\mu$), heat capacity ($Cv$), free energy ($G$), enthalpy ($H$), electronic spatial extent ($R^2$), internal energy at 298.15 K ($U$) and 0 K ($U_0$) , and zero point vibrational energy (ZPVE). 
The results are compared with NMP \cite{gilmer2017nmp}, SchNet \cite{schutt2017schnet}, Cormorant \cite{anderson2019cormorant}, L1Net \cite{miller2020l1net}, and LieConv \cite{finzi2020lieconv}.

Compared with the original backbone models, using the proposed positional encoding method (PosEnc) can improve the test error for both EGNN and DimeNet++, as listed in Table~\ref{table:eval}. The smoothness regularization technique (Smooth) can further increase the accuracy for several tasks.
%New lowest errors are achieved for the prediction of $\Delta \epsilon$, $\epsilon_\text{HOMO}$, $\epsilon_\text{LUMO}$ and $\mu$. 
We also conducted experiments using a 2-layer MLP with ReLU activation. For DimeNet++, replacing the manually-designed representation by an MLP generally increases test error compared to the original version. For EGNN, MLP increases error for $\epsilon_\text{HOMO}$ and $R^2$ and reduces error for the other tasks compared to the original version. The improvement however is not as significant as PosEnc.

\subsection{Regularized Embedding}

\begin{figure*}[t]
    \centering
    \includegraphics[width=0.8\textwidth]{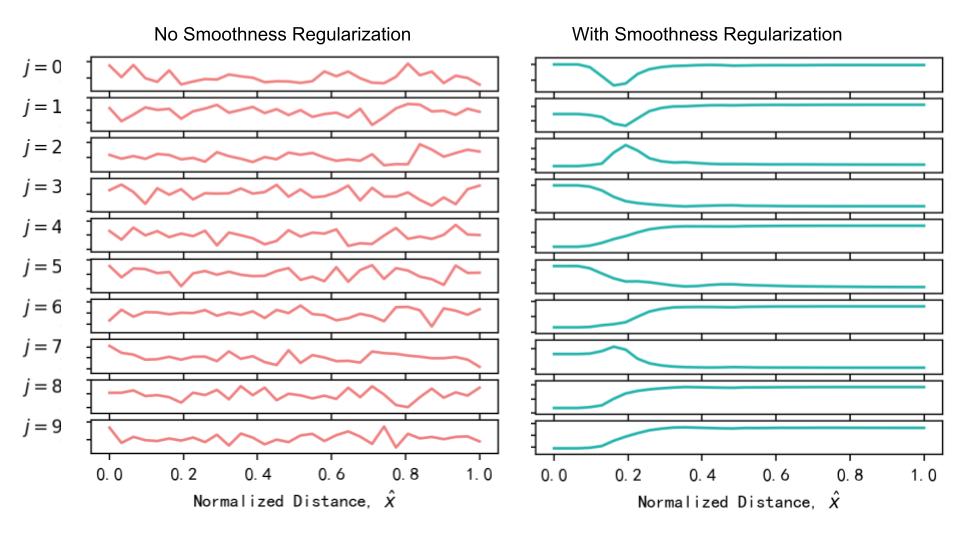}
    \caption{Values of the first 10 dimensions ($j=1\sim10$) of the distance embedding learned on QM9 $U_0$ task with EGNN backbone.}
    \label{fig:vis}
\end{figure*}

The smoothness regularization helps the learned embedding to be more smooth for easier interpretation.
The first 10 dimensions of the learned embedding are visualized in Figure~\ref{fig:vis} as a function of the normalized distance $x$ as \begin{align*}
\hat{x} &\equiv \frac{x - x_\text{min}} {x_\text{max} - x_\text{min}}.
\end{align*}

Without the smoothness regularization, the learned embedding appear to have no obvious patterns. The difference between the embeddings of two adjunct bins are large. In contrast, the embedding learned with smoothness regularization appear to be more smooth and show a simpler pattern. The embedding are generally nonlinear and non-monotonic. For the embedding learned for this particular task, $U_0$, we observe that the turning points concentrate at the small distance, and the embedding do not change much at the large distance.

\begin{figure*}[h]
    \centering
    \includegraphics[width=0.7\textwidth]{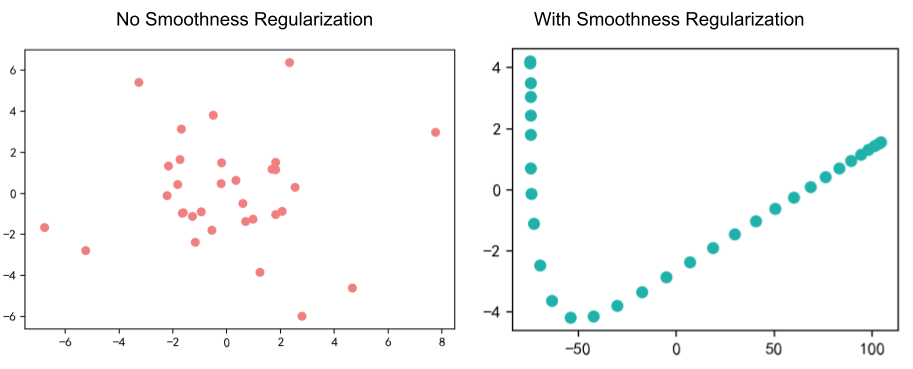}
    \caption{PCA visualization of the learned distance embedding on QM9 $U_0$ task with EGNN backbone model.}
    \label{fig:pca}
\end{figure*}

We conduct PCA analysis and visualize the learned embedding on a 2D space in Figure~\ref{fig:pca}. The embedding learned without regularization does not show a clear manifold, while the embedding learned with regularization forms a low-dimensional manifold. This 2-D manifold is not a straight line, consistent with the non-linearity and non-monotonicity observed above.

We further investigate the learned distance embedding in four aspects: non-linearity, non-monotonicity, diversity, and smoothness.

\begin{itemize}
    \item The non-linearity is measured based on Pearson's correlation, $\rho_\text{Pearson}$, between the distance and the embedding. $\text{Non-linearity} = 1 - \frac{1}{ls} \sum^{l}_{i=1} \sum^{s}_{j=1} \rho^2_\text{Pearson}(h_{ij}, x)$, where $l$ is the number of model layers and $h_{ij}$ is the $j$-th dimension of the embedding in the $i$-th layer. 

    \item The non-monotonicity is quantified based on Spearman's correlation, $\rho_\text{Spearman}$, between the distance and the embedding. $\text{Non-monotonicity} = 1 - \frac{1}{ls} \sum^{l}_{i=1} \sum^{s}_{j=1} \rho^2_\text{Spearman}(h_{ij}, x)$.
    
    \item The diversity indicates the average similarity between different dimensions of the learned embedding. A high diversity indicate that the embedding consists of diverse transforms.
    $\text{Diversity} = 1 - \frac{2}{ls(s-1)} \sum^{l}_{i=1} \sum^{s}_{j=1} \sum^{s}_{k=i+1} \rho^2_\text{Pearson}(h_{ij}, h_{ik})
    $.
    
    \item The smoothness is defined based on the smoothness loss
    $
        \text{Smoothness} = 1 - \mathcal{L}_\text{smooth}
    $.
\end{itemize}

\begin{table*}[!ht]
    \centering
    \small
    
    \begin{tabular}{p{0.17\textwidth} | p{0.02\textwidth} p{0.02\textwidth} p{0.035\textwidth} p{0.035\textwidth}  p{0.02\textwidth} p{0.02\textwidth} p{0.02\textwidth} p{0.02\textwidth} p{0.02\textwidth} p{0.02\textwidth} p{0.02\textwidth} p{0.05\textwidth} | p{0.09\textwidth}}
    
    \hline
      & \multicolumn{12}{c|}{Learned Distance Embedding} & Polynomial   \\
     & $\alpha$ & $\Delta\epsilon$ & $\epsilon_\text{HOMO}$ & $\epsilon_\text{LUMO}$ & $\mu$ & $C_v$ & $G$ & $H$ & $R^2$ & $U$ & $U_0$ & ZPVE & Baseline \\
    
    \hline
    Non-linearity &    .27 & .27 & .19 & .36 & .21 & .26 & .30 & .41 & .19 & .35 & .41 & .38 & .08\\
    Non-monotonicity & .19 & .18 & .13 & .24 & .15 & .23 & .22 & .29 & .18 & .33 & .33 & .28 & .00 \\
    Diversity &        .30 & .32 & .24 & .41 & .26 & .28 & .32 & .39 & .17 & .34 & .37 & .34 & .09 \\
    Smoothness &       .86 & .92 & .81 & .68 & .96 & .79 & .86 & .90 & .94 & .98 & .82 & .82 & .92\\     
    
    \hline
    \end{tabular}
    \caption{Characteristics of the distance embedding learned with smoothness regularization and a polynomial embedding.}
    \label{table:emb}
\end{table*}

These metrics are calculated for the embedding learned with smoothness regularization with EGNN model on QM9 dataset. For comparison, we consider a toy embedding made of three polynomial functions as its three dimensions:
$
     h_\text{polynomial}(x) = \{x, x^2, x^3 \}
     %\label{eq:poly}
$. As illustrated in Table~\ref{table:emb}, the learned embedding generally show higher non-monotocity and diversity compared to the polynomial embedding. The smoothness of the learned embedding are close to the polynomial embedding for some tasks. This indicates that, the proposed positional encoding method learn a embedding that contain multiple nonlinear or non-monotonic transformations, yet remain smooth. More physical-based interpretation of the learned embeddings are relegated to the appendix.

\subsection{Physics-based Interpretation }

In this appendix, we present some physic-based interpretation of learned embedding in various tasks from Section~\ref{sec-results}.2. Different tasks show different characteristics, as illustrated in Table~\ref{table:emb}. $U_0$ and $H$ show a high non-linearity, while $\mu$ and $R^2$ show a low non-linearity. In this section, we conduct a few case studies to investigate the connection between embedding characteristics and the physical nature of the corresponding task.

\subsubsection{Short vs. Long-Distance Physics}
\begin{figure}[h]
    \centering
    \includegraphics[width=0.43\textwidth]{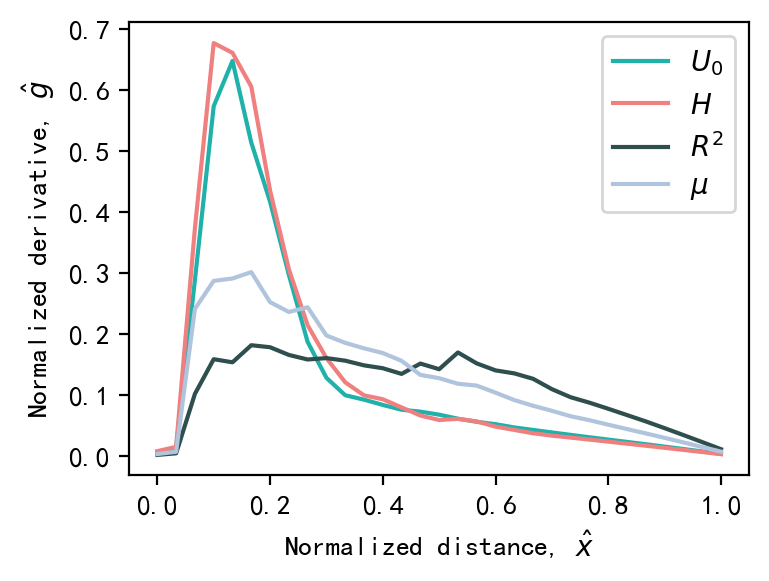}
    \caption{Distribution of derivative over distance for different tasks.}
    \label{fig:derivative}
\end{figure}

The smoothness regularization encourage the embedding to only change with $x$ when such change increases the accuracy. By analyzing how much the learned distance embedding changes with distance, we know which region of distance the model learned to focus on. 
We quantify the change by the normalized derivative.
\begin{align*}
     \hat{g}(x) &\equiv  \frac{\text{abs}(g(x))} {\text{std}(g(x))}.
\end{align*}
As illustrated in Figure~\ref{fig:derivative}, for both $U_0$ and $H$, $\hat{g}(x)$ is large at small $\hat{x}$, and then quickly drop to a small value close to zero as $\hat{x}$ increases. This indicates that the embedding values do not change much when the distance is greater than certain "cutoff" value. This is consistent with the physical nature of the problem. $U_0$ and $H$ of the system considered in QM9 are generally dominated by short-range interactions. There is no significant long-range forces such as electrostatic force in these systems. 

In contrast, for both $\mu$ and $R^2$, $\hat{g}(x)$ show a much flatter profile. This indicates that both long distance and short distance are important. This observation is consistent with the physical natural of the tasks. 
% https://byjus.com/chemistry/dipole-moment/
Dipole moment ($\mu$) is roughly the charges timed by the distance. By definition it is a quantity sensitive to both short and long distance. Therefore the embedding learned for this task has different value at different distance, as shown by the non-zero $\hat{g}(x)$ over full $\hat{x}$ range.
% http://pollux.chem.umn.edu/4502/3502_lecture_32.pdf
Electronic spatial extent ($R^2$) reports how far out the electronic density extends with significant probability. This roughly measure the molecule size and is obviously not just sensitive to short interatomic distance. %Similar to the case of $\mu$, the physical nature of this task requires the learned distance embedding to not converge as distance increases.

These case studies illustrate that the learned embedding reflect the physical dependence on short and long distance for different tasks.

\subsubsection{Monotonic vs. Non-monotonic Physical Dependence}

%\input{figures/turning_point}

%The embedding learned with the smoothness regularization shows a simpler pattern. As shown in Figure~\ref{fig:vis}, for most dimension, there is a clear turning point, which separate two regions of opposite monotonic curves. We analyze the non-monotonicity of the learned embedding by investigating the distribution of the turning points. For a given dimension $j$ and layer $i$, we define an $x$ is a turning point if $h_{i,j}(x)$ is greater than the embedding at both left and right side of $x$, or it is smaller than both sides. We visualize the average value of this 0/1 classification.

%As illustrated in Figure~\ref{fig:turning_point}, for $U_0$ task, the turning points appear frequently at short distance region. In comparison, the turning points are not clearly observed for $R2^2$. 

The non-monotonicity of the learned distance embedding illustrated in Table~\ref{table:emb} is rooted in the physical nature of the tasks. 

For physical quantity governed by non-monotonic dependence on distance, the corresponding learned distance embedding shows a high non-monotonicity. 
The internal energy is connected to the potential forces acting on the atoms. The force can be non-monotonic w.r.t. distance. Short distance is often dominated by repulsive forces while attractive forces are more significant at larger distance. Consistent with this non-monotonic physics, The embeddings learned for $U_0$ and $U_0$ have the highest non-monotonicity score, 0.33, listed in Table~\ref{table:emb}.

In contrast, if there is no strong non-monotonic physical dependence on distance, the learned distance embedding has a low non-monotonicity score.
For instance, $R^2$ roughly measure the molecule size, as mentioned above, and is monotonic with distance. The distance embedding learned for $R^2$ consequently shows the a relatively low non-monotonicity score, 0.18.

\subsubsection{Similar Physics have Similar Distance Embedding}

\begin{figure*}[ht]
    \centering
    \includegraphics[width=0.5\textwidth]{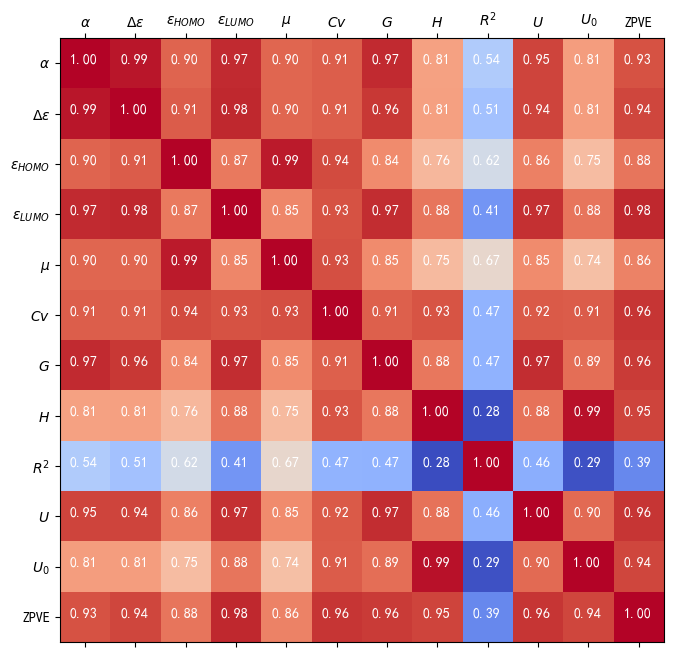}
    \caption{Pairwise similarity of the learned distance embedding.}
    \label{fig:mat}
\end{figure*}
The tasks can be closely related by their similar physical nature. Some relation is obvious. Both $U$ and $U_0$ are internal energy, and there is a high correlation between their values in the QM9 dataset.
Some physical connection is not straightforward. For example, the energy gap $\Delta \epsilon$ have been used to predict the polarizability $\alpha$ \cite{ravindra1979alphagap, reddy1996alphagap}, but such relation can not be simply observed in the original data. The Pearson correlation between $\alpha$ and $\Delta \epsilon$ is close to zero in QM9. Can the learned distance embedding capture such physical relation?

We hypothesize that if two tasks are governed by the similar set of physical laws, the embedding learned for these two tasks should share certain similarity. We quantify the similarity between the embedding for task $a$ and task $b$ based on Pearson correlation.
\begin{align*}
     \text{Similarity} = \frac{1}{ls^2} \sum^{l}_{i=1} \sum^{s}_{j=1} \sum^{s}_{k=1} \rho^2_\text{Pearson}(h^{\text{task}~a}_{i,j}, h^{\text{task}~b}_{i,k}.)
\end{align*}
The pairwise similarity is visualized in Figure~\ref{fig:mat}. The energies $U$, $U_0$, $H$, $G$ and ZPVE show a high similarity ($\geq 0.88$) between each other.
Another group of physical quantities, $\alpha$, $\mu$, and $\Delta \epsilon$ are closely related to the reactivity of the molecules. The pairwise embedding similarity for this group is high ($\geq 0.90$).
In comparison, the similarity cross these two groups are relatively low. 
This observation indicates that tasks of the similar physical nature have the similar embedding.

% TODO: it seems the HOMO LUMO gap 
%Both $\mu$ and $\alpha$ are related to the dipole moment and their 

%additional_moment = polarity * electric
% https://chemistry.stackexchange.com/questions/51292/difference-between-polarizability-and-dipole-moment

\section{Conclusion}
\label{sec-conclusion}

We propose and demonstrate a regularized positional encoding method for molecular properties prediction tasks. 
As a plug-in module, the proposed method improves the accuracy of three model architecures, over variety molecular property and force field prediction tasks.
The novel smoothness regularization technique encourages the learned embedding to form a simpler low-dimensional manifold for easier physics-based interpretation. Key characteristics of the embedding are connected to the physical nature of the tasks. Tasks of similar physics have the similar learned embedding. 


\clearpage
\newpage

{
\small
\bibliographystyle{plainnat}
\bibliography{ref}
}

\clearpage
\newpage
%\input{sections/checklist}

%\clearpage
%\newpage
%\appendix
%\input{sections/appendix}
%\input{sections/4.4.physics}

\end{document}